\documentclass{article}
\usepackage{spconf,amsmath,graphicx}

\usepackage{xcolor}

\title{Interpreting Galaxy Deblender GAN from the Discriminator's Perspective}
\name{Heyi Li$^{\star}$ \qquad Yuewei Lin$^{\dagger}$ \qquad Klaus Mueller$^{\star}$ \qquad Wei Xu$^{\dagger}$}
\address{$^{\star}$ Department of Computer Science, Stony Brook University, United States \\ $^{\dagger}$ Computer Science Initiative, Brookhaven National Laboratory, United States}
\begin{document}
%
\maketitle
\begin{abstract}
Generative adversarial networks~(GANs) are well known for their unsupervised learning capabilities. A recent success in the field of astronomy is deblending two overlapping galaxy images via a branched GAN model. However, it remains a significant challenge to comprehend how the network works, which is particularly difficult for non-expert users. This research focuses on behaviors of one of the network's major components, the Discriminator, which plays a vital role but is often overlooked, Specifically, we enhance the Layer-wise Relevance Propagation~(LRP) scheme to generate a heatmap-based visualization. We call this technique Polarized-LRP and it consists of two parts i.e. positive contribution heatmaps for ground truth images and negative contribution heatmaps for generated images. Using the Galaxy Zoo dataset we demonstrate that our method clearly reveals attention areas of the Discriminator when differentiating generated galaxy images from ground truth images. To connect the Discriminator's impact on the Generator, we visualize the gradual changes of the Generator across the training process. An interesting result we have achieved there is the detection of a problematic data augmentation procedure that would else have remained hidden. We find that our proposed method serves as a useful visual analytical tool for a deeper understanding of GAN models. 
\end{abstract}
\section{Introduction}
\label{sec:intro}

Astronomical researchers routinely assume the strict isolation of the targeted celestial body and so their objective is simplified into evaluating the properties of a single object. However, galactic overlapping is ubiquitous in current surveys due to projection effects and source interactions. This introduces bias to multiple physical traits such as photometric redshifts and weak lensing at levels beyond requirements. With the arrival of the next generation of ground-based galaxy surveys such as the Large Synoptic Survey Telescope~(LSST)~\cite{ivezic2019lsst} which is expected to begin operation in 2023, this issue becomes more urgent. Specifically, the increase of both depth and sensitivity will cause the number of blended galaxy images to grow exponentially. Dawson~\cite{dawson2015ellipticity} predicts that nearly $50\%$ of galaxies captured in LSST images encounter overlapping with a $3"$ center-to-center distance. This leads to immense quantities of imaging data warranted as unusable. According to the estimation in \cite{reiman2019deblending}, up to $200$ Million galaxy images could be discarded each year if the blending issue is not effectively addressed throughout the ten year period of the LSST survey. However, the task of galaxy deblending remains a open problem in the field of astronomy and no gold standard solution exists in the processing pipeline. 

Since Goodfellow~\cite{goodfellow2014generative} first proposed the generative adversarial network~(GAN) model, it has achieved state-of-the-art performance in many computer vision applications, especially in face generation \cite{chen2017occlusion,antipov2017face,karras2019style} and image manipulation \cite{isola2017image, zhu2017unpaired}. Many GAN variants \cite{arjovsky2017wasserstein,li2018fast} have been proposed to improve the training stability and boost the convergence speed. Recently, the GAN model \cite{reiman2019deblending} has been applied in solving the galaxy deblending problem and has yielded promising results in the conspicuous (confirmed blends) class of blends. However, a conceptual understanding of GAN models is still largely lacking, which makes building and training GAN models extremely demanding for non-expert users in the astronomy society. This prohibits wide utilization of GAN models and potentially prevents GAN models from reaching optimum performance. More importantly, the lack of interpretation directly results in the lack of trust in images generated by GAN models. 

During our discussions with domain scientists, we noticed two facts: (1) a visual explanation can help them understand model behaviour without machine learning expertise, and (2) the behavior of the Discriminator is most perplexing to astronomers. Different visualization algorithms have been proposed to increase the interpretability of convolutional neural networks~(CNNs). Class activation mapping~(CAM) based methods \cite{zhou2016learning,selvaraju2017grad} directly use the activation of the last convolutional layer to infer the downsampled relevance of the input pixels. But such methods are only applicable to specific architectures which use the average pooling layer. The layer-wise relevance propagation~(LRP) algorithm \cite{montavon2019layer} is proposed to address this issue. For each image, LRP propagates the classification score backward through the model and calculates relevance intensities over all pixels. Although successful in interpreting discriminative classifiers \cite{li2019beyond}, the LRP algorithm does not cover network structures like GAN models. Among the limited works explaining generative network models, Liu \cite{liu2017analyzing} designed a GUI interface to display connections between neurons of neighboring layers in a GAN model. Unfortunately, their tool is intended only for machine learning experts and hence not supportive for non-domain researchers. Most recently, Bau \cite{bau2019gan} came up with a dissecting framework which examines the causal relationship between network units and object concepts. Likewise, the relationship between face aging and model parameters is examined in \cite{genovese2019towards} but the Discriminator is completely omitted in their works. Although not used to generate images during the inference stage, the Discriminator significantly affects the performance of the Generator, which is important to investigate. 

Therefore, we propose to extend the LRP framework to bridge this gap. Our Polarized-LRP propagates the single probability value given by the Discriminator backwards, during which it calculates positive contributions for ground truth images and negative contributions for generated images separately. By comparing relevance maps for the same input at different iterations during the training stage, our method clearly reveals the gradual changes of the Generator in response to the direct feedback from the Discriminator. Moreover, we demonstrate the effectiveness of our method by uncovering a problematic step in data augmentation which was previously unknown to astronomers. To the best of our knowledge, our Polarized-LRP is the first method in the literature which can effectively visualize the behavior of the Discriminator and its impact on the Generator. 

The remainder of this paper is organized as follows. A thorough explanation of our new visualization scheme is included in Section \ref{sec:method}. Extensive experiments are conducted to prove the effectiveness of our method, which are described in Section \ref{sec:experiments}. We present our conclusions and discussions of future work in Section \ref{sec:conclusion}. 

\section{Our Visualization Method}
\label{sec:method}

\subsection{Polarized-LRP}
\label{ssec:polarized-lrp}

As mentioned in Section \ref{sec:intro}, the LRP algorithm applies only to discriminative models. The root of this limitation lies in the structure of the relevance input. The classifier's output consists of the predicted probability for each class. All elements in this vector are adjusted to zero except the highest one which represents the target class. This altered vector then serves as the relevance input. In this way, only neurons connected to this element are activated during the propagation. The Discriminator, on the other hand, only returns one probability value. Applying the LRP algorithm directly is essentially assuming that there is only one class, which renders all output heatmaps meaningless. 

To address this issue, our Polarized-LRP computes two relevance maps from the same probability value, one positive and one negative. The positive relevance map displays only positive contributions from pixels in the input image to the probability value, while the negative relevance map shows only negative contributions. If one image is classified as fake by the Discriminator, then the negative contributions from input pixels dominate and thus decrease the probability score. In this case, the negative relevance map represents and conveys the decision making of the Discriminator. The opposite is true if the image is classified as real. By polarizing the relevance into positive and negative, our algorithm creates two "virtual" classes from the Discriminator's output probability. 

Our Polarized-LRP follows the conservation rule which guarantees the summation of relevance $R$ propagated through all neurons $d$ in each layer $L$ of the model to remain constant. 

\begin{equation} 
\label{eq:eq1}
\sum_{d \in V_L} R_d^{(L)} = \cdots = \sum_{d \in V_l} R_d^{(l)} = \cdots = \sum_{d \in V_1} R_d^{(1)}
\end{equation}

The conservation rule also assures that the inflow of relevance to one neuron matches the outflow of relevance from that same neuron, which is captured in Equation \ref{eq:eq2}. Neurons such as $m$ and $n$ in layer $l+1$ are connected to neuron $d$ which in turn is connected to neurons such as $p$ and $q$ in layer $l-1$. 

\begin{equation} 
\label{eq:eq2}
    \begin{aligned}
    R_d^{(l)} 
    = R_{m \rightarrow d}^{(l+1) \rightarrow (l)} + \cdots + R_{n \rightarrow d}^{(l+1) \rightarrow (l)} \\
    = R_{d \rightarrow p}^{(l) \rightarrow (l-1)} + \cdots + R_{d \rightarrow q}^{(l) \rightarrow (l-1)}
    \end{aligned}
\end{equation}

Here $R_{j \rightarrow i}^{(l+1) \rightarrow (l)}$ represents the contribution from neuron $j$ at layer $l+1$ to neuron $i$ at layer $l$. It is computed using the propagation procedure in Equation \ref{eq:eq3}. 

\begin{equation} 
\label{eq:eq3}
R_{j \rightarrow i}^{(l+1) \rightarrow (l)} = 
\begin{cases}
\frac{[w_{ij} x_i]^{+}}{\sum_k [w_{kj} x_k]^{+} + b_k^{+}} R_j^{(l+1)}, & \text{if classified Real} \\
\frac{[w_{ij} x_i]^{-}}{\sum_k [w_{kj} x_k]^{-} + b_k^{-}} R_j^{(l+1)}, & \text{if classified Fake}
\end{cases}
\end{equation}

The weights and biases are denoted by $w_{ij}$ and $b_k$ respectively. ${[]}^{+}$ and ${[]}^{-}$ represent value truncation at zero. 

\subsection{Demonstration Examples}
\label{ssec:examples}

\begin{figure}[htb]

\begin{minipage}[b]{0.48\linewidth}
  \centering
  \centerline{\includegraphics[width=4.0cm]{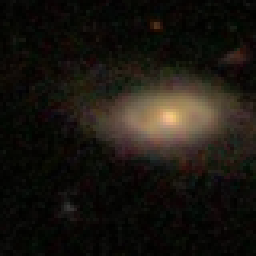}}
  \centerline{(a) ground truth image}\medskip
\end{minipage}
\hfill
\begin{minipage}[b]{0.48\linewidth}
  \centering
  \centerline{\includegraphics[width=4.0cm]{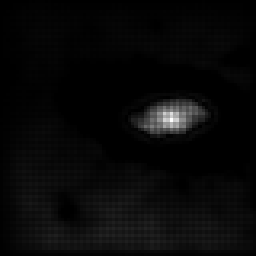}}
  \centerline{(b) positive relevance map}\medskip
\end{minipage}
\begin{minipage}[b]{0.48\linewidth}
  \centering
  \centerline{\includegraphics[width=4.0cm]{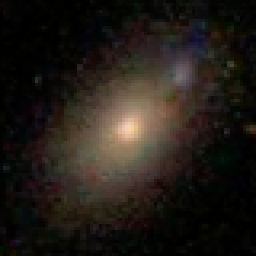}}
  \centerline{(c) generated image}\medskip
\end{minipage}
\hfill
\begin{minipage}[b]{0.48\linewidth}
  \centering
  \centerline{\includegraphics[width=4.0cm]{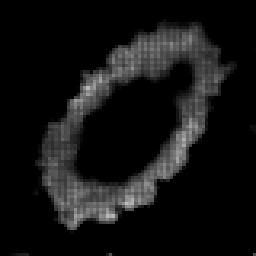}}
  \centerline{(d) negative relevance map}\medskip
\end{minipage}
\caption{The first row includes an example of the positive relevance map and the second row contains an example of the negative relevance map. Image are enlarged to $256 \times 256$ for better visualization.}
\label{fig:fig1}
\end{figure}

Here we present two cases as examples of demonstration for our proposed Polarized-LRP. The first row in Figure \ref{fig:fig1} shows the positive relevance map for a ground truth image. From the map on the right, we can see that the Discriminator focuses on the interior of the galaxy ellipse. Pixels in the attention area make strong positive contributions to the probability score, which explains why the Discriminator classifies this image as real. The second row in  Figure \ref{fig:fig1} exhibits the negative relevance map for a generated image. The map indicates that the Discriminator makes its decision based on pixels encircling the central area, which is the most noticeable area in the image. This is consistent with the visual inspection between the real image and the generated image by a domain expert.

\section{Experiments and Discussions}
\label{sec:experiments}

To explain the role of the Discriminator to astronomers, we first select one successful training instance of the galaxy deblender GAN model and compare relevance maps of the same input at different training iterations. Next, during our analysis of failed training attempts, we discover an unusual pattern which leads to the successful diagnosis of an erroneous data augmentation procedure.

\subsection{Galaxy Deblender GAN Model Replication}
\label{ssec:replication}

Since no pre-trained network is publicly available, we re-train the galaxy deblender GAN from scratch. All steps mentioned in \cite{reiman2019deblending} are strictly followed. Training and testing datasets are re-generated using the raw galaxy images in the Kaggle Galaxy Zoo classification challenge \cite{lintott2010galaxy}. The GAN model is trained using the same learning rate setting and is updated for the same number of iterations. One single Tesla V100 graphics card was used for the training. Table \ref{tab:tab1} shows the reported peak noise-to-signal ratio~(PSNR) value and structural similarity index~(SSIM) value along with ours. Although our values are slightly lower than their reported ones, they are within a reasonable shift range. 

\begin{table}
\begin{center}
\begin{tabular}{|c|c|c|}
\hline
Mean & PSNR(dB) & SSIM \\
\hline\hline
Reported & 34.61 & 0.92 \\
Replicated & 33.47 & 0.89 \\
\hline
\end{tabular}
\end{center}
\caption{Mean values of PSNR and SSIM}
\label{tab:tab1}
\end{table}

\begin{figure}[htb]

\begin{minipage}[b]{0.3\linewidth}
  \centering
  \centerline{\includegraphics[width=2.8cm]{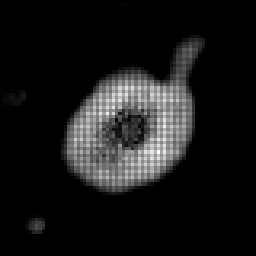}}
  \centerline{(a) relevance map}\medskip
\end{minipage}
\hfill
\begin{minipage}[b]{0.3\linewidth}
  \centering
  \centerline{\includegraphics[width=2.8cm]{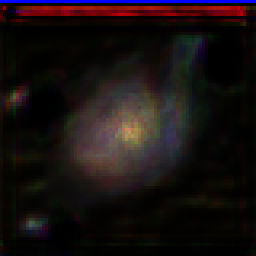}}
  \centerline{(b) early stage}\medskip
\end{minipage}
\hfill
\begin{minipage}[b]{0.3\linewidth}
  \centering
  \centerline{\includegraphics[width=2.8cm]{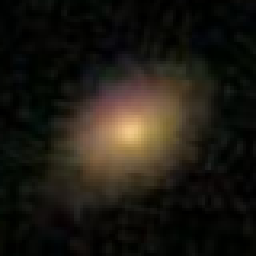}}
  \centerline{(c) ground truth}\medskip
\end{minipage}
\begin{minipage}[b]{0.3\linewidth}
  \centering
  \centerline{\includegraphics[width=2.8cm]{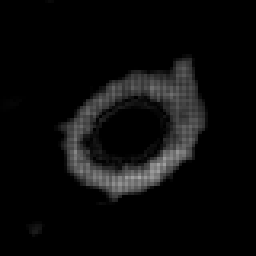}}
  \centerline{(d) relevance map}\medskip
\end{minipage}
\hfill
\begin{minipage}[b]{0.3\linewidth}
  \centering
  \centerline{\includegraphics[width=2.8cm]{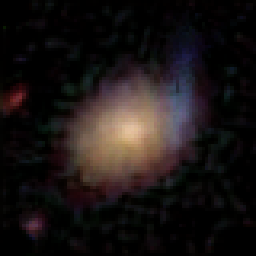}}
  \centerline{(e) middle stage}\medskip
\end{minipage}
\hfill
\begin{minipage}[b]{0.3\linewidth}
  \centering
  \centerline{\includegraphics[width=2.8cm]{figures/1_gt.png}}
  \centerline{(f) ground truth}\medskip
\end{minipage}
\begin{minipage}[b]{0.3\linewidth}
  \centering
  \centerline{\includegraphics[width=2.8cm]{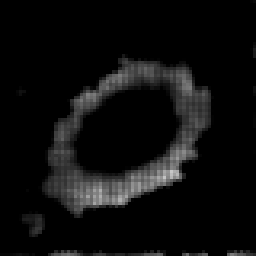}}
  \centerline{(g) relevance map}\medskip
\end{minipage}
\hfill
\begin{minipage}[b]{0.3\linewidth}
  \centering
  \centerline{\includegraphics[width=2.8cm]{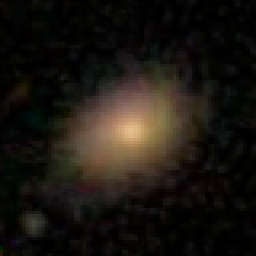}}
  \centerline{(h) final stage}\medskip
\end{minipage}
\hfill
\begin{minipage}[b]{0.3\linewidth}
  \centering
  \centerline{\includegraphics[width=2.8cm]{figures/1_gt.png}}
  \centerline{(i) ground truth}\medskip
\end{minipage}
\caption{Each row represents one different stage during training. Images are enlarged to $256 \times 256$ for better visualization.}
\label{fig:fig2}
\end{figure}

\subsection{Training Understanding}
\label{ssec:understanding}

An example of a high-quality generation is shown in Figure \ref{fig:fig2}. Three iterations are selected correspondingly at an early stage, at an approximate mid-point, and near the end when the model converges. For the three generated images, the Discriminator gave a score of $0.001$, $0.001$, $0.003$. We plot their relevance maps as to indicate why they are identified as fake. From the central image in the first row, we can see that the Generator only manages to replicate the inner bright spot. In the Discriminator's relevance map, this corresponds to the small black hole in the middle. Furthermore, our relevance map on the left clearly reveals that the low probability score by the Discriminator is mostly due to the unrealistic-looking pixels in the surrounding areas. This information is then passed on to the Generator as the adversarial loss penalty. As is shown in the images in the subsequent rows, the white ring in our relevance map grows thinner and darker. Along with the expansion of the black hole (meaning the confidence area in the center of galaxy enlarges), the generated image slowly transforms towards the ground truth image. One interesting finding is that the Generator learns the glaring spot first and then incrementally apprehends the surroundings. This is comprehensible because the Generator is first trained with only the style loss from VGG19. During this so-called "burn-in" period, features such as salient spots are expected to be grasped by the Generator. The benefit of the style loss during the "burn-in" period is easily visualized in our relevance map. How the Generator changes during training under the guidance of the Discriminator is also revealed.

\begin{figure}[htb]

\begin{minipage}[b]{0.3\linewidth}
  \centering
  \centerline{\includegraphics[width=2.8cm]{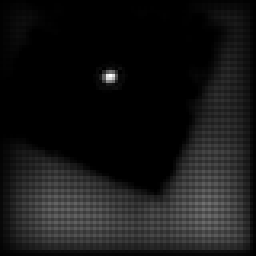}}
  \centerline{(a) relevance map}\medskip
\end{minipage}
\hfill
\begin{minipage}[b]{0.3\linewidth}
  \centering
  \centerline{\includegraphics[width=2.8cm]{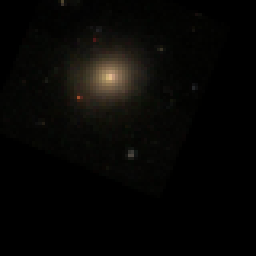}}
  \centerline{(b) ground truth}\medskip
\end{minipage}
\hfill
\begin{minipage}[b]{0.3\linewidth}
  \centering
  \centerline{\includegraphics[width=2.8cm]{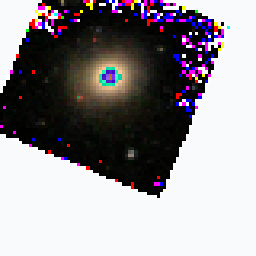}}
  \centerline{(c) contrast increased}\medskip
\end{minipage}
\caption{The "phantom boundary" becomes apparent when the contrast of the ground truth image is increased. Images are enlarged to $256 \times 256$ for better visualization.}
\label{fig:fig3}
\end{figure}

\subsection{Model Debugging}
\label{ssec:debugging}

While analyzing the galaxy deblender GAN using our method, we noticed a strange phenomenon consistently appearing in the positive relevance maps. Consider Figure \ref{fig:fig3} where the boundary of a rectangular shape is shown in the positive relevance map of the ground truth image. This shape only appears in the positive relevance maps which is an important revelation as it tells us that the Discriminator picks up features from the ground truth images which are hidden from our sight. A partial decision was mistakenly made from the image background without any footing in domain knowledge.

\begin{figure}[htb]
\begin{minipage}[b]{1.0\linewidth}
  \centering
  \centerline{\includegraphics[width=8.5cm]{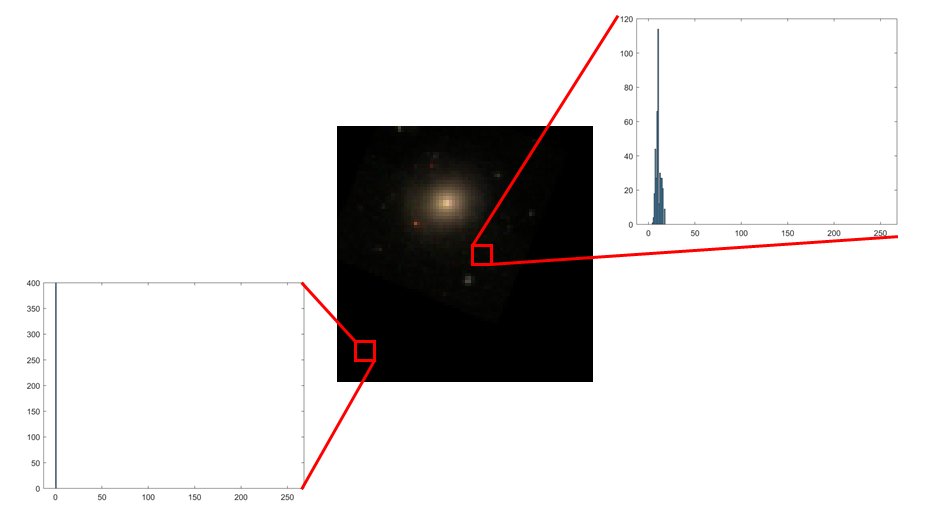}}
\end{minipage}
\caption{Two $20 \times 20$ background regions randomly selected from the ground truth image. While they look alike visually, these two regions have very different histograms. The ground truth image was enlarged ($256 \times 256$) for better visualization.}
\label{fig:fig4}
\end{figure}

Further investigation revealed that this "phantom boundary" was introduced in the data preparation stage. One image out of each blended pair was randomly perturbed by flipping, rotation, displacement, and scaling. Then, after these operations, all missing pixels in the newly created image were filled with zeros. However, this padded true black background diverges from the near black background of the galaxy although the two seem quite alike with visual inspection. Figure \ref{fig:fig4} shows the histogram of two different $20 \times 20$ regions in the ground truth image, one from the galaxy background and the other from the manually padded background.

This problem had a large impact as it crippled our galaxy deblender GAN model from reaching its optimum performance. Instead of capturing features of real celestial bodies, the Discriminator learned a much simpler strategy to manipulate the equilibrium system utilizing the "phantom boundary". No matter how realistic the generated images were by both visual inspection and SSIM metrics, the Discriminator could easily differentiate them as long as the "phantom boundary" was absent. While zero-padding is a frequently used technique in image processing, many non-domain experts are unaware of its shortcoming. Fortunately, with the help of our proposed algorithm this problems could be detected. It was eventually resolved by replacing the zero-padding with a random noise distribution obtained from physics statistics. 

\section{Conclusion}
\label{sec:conclusion}

Despite the many successes of GAN models they are still difficult to understand. We propose a Polarized-LRP technique that allows interpretation of the galaxy deblender GAN from the perspective of the Discriminator. By visualizing positive and negative contributions separately, our algorithm successfully reveals how the Discriminator detects generated images and how it affects the Generator during training. 
Our method was immediately useful in uncovering a hidden mistake in the data preparation stage of our collaborators' galaxy images. Although designed for the galaxy deblending problem, Polarized-LRP is not restricted to this network. We plan to apply our method to interpret well-established GAN models such as StyleGAN and others. We also plan to extend LRP to the Generator as well to provide a complete understanding of both GAN components. Finally, we will design a visual analytics system to also facilitate direct user interaction.

\bibliographystyle{ieee}
\bibliography{Template}

\end{document}